\documentclass{article} 
\usepackage{iclr2021_conference,times}

\usepackage{graphicx}
\usepackage{subfigure}
\usepackage{float}
\usepackage{amsmath}
\usepackage{amssymb}
\usepackage{multirow}
\usepackage{booktabs}
\usepackage{url}
\usepackage{array}
\usepackage{enumitem}
\usepackage{algorithm}
\usepackage{algorithmic}
\usepackage{pifont}
\usepackage{bm}
\usepackage{CJKutf8}
\usepackage{CJK,CJKnumb} 
\usepackage{makecell}

\usepackage{hyperref}
\usepackage{url}
\iclrfinalcopy

\title{PERT: Pre-training BERT with Permuted Language Model}

\author{Yiming Cui$^{1,2,3}$, Ziqing Yang$^{2,3}$, Ting Liu$^1$ \\
{$^1$Research Center for Social Computing and Information Retrieval (SCIR), School of Computer} \\
{Science \& Technology, Harbin Institute of Technology, Harbin, China} \\
{$^2$State Key Laboratory of Cognitive Intelligence, iFLYTEK Research, Beijing, China} \\
{$^3$Joint Laboratory of HIT and iFLYTEK Research (HFL), Beijing, China} \\
{$^1$\tt\{ymcui,tliu\}@ir.hit.edu.cn} \\
{$^{2,3}$\tt\{ymcui,zqyang5\}@iflytek.com} \\  
}

\begin{document}
\begin{CJK*}{UTF8}{gkai}

\maketitle

\begin{abstract}
Pre-trained Language Models (PLMs) have been widely used in various natural language processing (NLP) tasks, owing to their powerful text representations trained on large-scale corpora. 
In this paper, we propose a new PLM called PERT for natural language understanding (NLU).
PERT is an auto-encoding model (like BERT) trained with Permuted Language Model (PerLM). 
The formulation of the proposed PerLM is straightforward.
We permute a proportion of the input text, and the training objective is to predict the position of the original token.
Moreover, we also apply whole word masking and N-gram masking to improve the performance of PERT.
We carried out extensive experiments on both Chinese and English NLU benchmarks.
The experimental results show that PERT can bring improvements over various comparable baselines on some of the tasks, while others are not.
These results indicate that developing more diverse pre-training tasks is possible instead of masked language model variants.
Several quantitative studies are carried out to better understand PERT, which might help design PLMs in the future.\footnote{Resources are available: \url{https://github.com/ymcui/PERT}}
\end{abstract}

\section{Introduction}
Pre-trained Language Models (PLMs) have shown excellent performance on various natural language processing (NLP) tasks. 
Usually, PLMs are classified into two categories based on their training protocols: auto-encoding and auto-regressive.
The representative work of auto-encoding PLM is the Bidirectional Encoder Representations from Transformers (BERT) \citep{devlin-etal-2019-bert}, which models the input text through deep transformer layers \citep{vaswani2017attention}, creating deep contextualized representations.
The representative work of auto-regressive PLM is the Generative Pre-training (GPT) model \citep{radford2018improving}. 
In this paper, we mainly focus on the auto-encoding PLMs.

The dominant pre-training task for auto-encoding PLM is the masked language model (MLM). 
The MLM pre-training task replaces a few input tokens with masking tokens (i.e., {\tt [MASK]}), and the objective is to recover these tokens in the vocabulary space.
The formulation of MLM is pretty simple, but it can model the contextual features around the masked token, which is quite similar to the continuous bag-of-words (CBOW) in word2vec \citep{mikolov-etal-2013}.
Based on MLM pre-training task, there are also a few variants proposed to further enhance its performance, such as whole word masking \citep{devlin-etal-2019-bert,sun2019ernie,chinese-bert-wwm}, N-gram masking \citep{devlin-etal-2019-bert,joshi2019spanbert,cui-etal-2020-revisiting}, etc.
Following MLM pre-training scheme, various PLMs are proposed, such as ERNIE \citep{sun2019ernie}, RoBERTa \citep{liu2019roberta}, ALBERT \citep{lan2019albert}, ELECTRA \citep{clark2020electra}, MacBERT \citep{chinese-bert-wwm}, etc.
\begin{table*}[h]
\small
\begin{center}
\begin{tabular}{l l l }
\toprule
& \bf Input & \bf Output \\
\midrule
\bf Original Text & \text{研究表明这一句话的顺序并不影响阅读。} & - \\
\midrule
\bf WordPiece &  \text{研~究~表~明~这~一~句~话~的~顺~序~并~不~影~响~阅~读~。} & - \\
\midrule
\bf BERT & \text{研~究~表~明~这~一~句~{\bf [M]}~的~顺~{\bf [M]}~并~不~{\bf [M]}~响~阅~读~。} & \makecell[lt]{Pos$_7$ $\to$ 话 \\ Pos$_{10}$ $\to$ 序 \\ Pos$_{13}$ $\to$ 影} \\
\midrule
\bf PERT & \text{研~究~{\bf 明}~{\bf 表}~这~一~句~话~的~顺~序~并~不~{\bf 响}~{\bf 影}~阅~读~。} & \makecell[lt]{Pos$_2$ $\to$ Pos$_3$ \\ Pos$_3$ $\to$ Pos$_2$ \\ Pos$_{13}$ $\to$ Pos$_{14}$ \\ Pos$_{14}$ $\to$ Pos$_{13}$} \\
\bottomrule
\end{tabular}
\end{center}
\caption{\label{example} Comparisons of input and output for BERT and PERT. Pos: Position.}
\end{table*}

However, there comes with a natural question: {\bf\em Can we use pre-training task other than MLM?}
To deal with this question, in this paper, we aim to explore a pre-training task that is not derived from MLM.
The original motivation behind our approach is quite interesting.
There are many sayings like ``Permuting several Chinese characters does not affect your reading that much''. 
A vivid illustration of this phenomenon is depicted in Figure \ref{example}.
As we can see, with a first glimpse, we might not notice that some words in the sentence are disordered, but we can still grasp the central meaning of the sentence.
This phenomenon makes us curious whether we can model the contextual representation via permuted sentences.
To investigate this question, we propose a new pre-training task, called permuted language model (PerLM). 
The proposed PerLM tries to recover the word orders from a disordered sentence, and the objective is to predict the position of the original word.
We pre-train both Chinese and English PERT to examine their effectiveness.
Extensive experiments are conducted on both Chinese and English NLP datasets, ranging from sentence-level to document-level, such as machine reading comprehension, text classification, etc.
The results show that the proposed PERT can give improvements on a few tasks.
While in the meantime, we also discover their deficiencies in others.
The contributions of this paper are listed as follows.
\begin{itemize}
	\item We propose a non-MLM-like pre-training task, called permuted language model, which tries to recover the shuffled input text into the right order.
	\item Experimental results show both positive and negative results, and further analyses might be helpful in future research.
	\item Pre-trained Chinese and English PERT models are made publicly available.
\end{itemize}

\section{Related Work}
In this section, we revisit the techniques of the representative pre-trained language models in the recent natural language processing field.
Text representations have made significant progress in recent years, after the emergence of the pre-trained language models.
The pre-trained language model utilizes large-scale text corpora and unsupervised (or self-supervised) learning algorithms to extract text semantics in a continuous space.
This paper mainly focuses on the pre-trained language model for natural language understanding (NLU).
A list of such PLMs are listed in Table \ref{model-comparison}, including the newly proposed PERT.
Next, we briefly introduce these models.
\begin{table*}[h]
\small
\begin{center}
\begin{tabular}{l c c c c c}
\toprule
& \bf Type & \bf Embeddings & \bf Masking & \bf LM Task & \bf Paired Task \\
\midrule
GPT \citep{radford2018improving}	& AR & T/S/P & - & LM & - \\
BERT \citep{devlin-etal-2019-bert}	& AE & T/S/P & T & MLM & NSP \\
ERNIE \citep{sun2019ernie} 			& AE & T/S/P & T/E/Ph & MLM & NSP \\
XLNet \citep{yang2019xlnet}			& AR & T/S/P & - & PLM & - \\
RoBERTa \citep{liu2019roberta} 		& AE & T/S/P & T & MLM & - \\
ALBERT \citep{lan2019albert} 		& AE & T/S/P & T & MLM & SOP \\
ELECTRA \citep{clark2020electra}	& AE & T/S/P & T & Gen-Dis & - \\
MacBERT \citep{chinese-bert-wwm}	& AE & T/S/P & WWM/NM & Mac & SOP \\
\midrule
\bf PERT 							& AE & T/S/P & WWM/NM & PerLM & - \\
\bottomrule
\end{tabular}
\end{center}
\caption{\label{model-comparison} Comparisons of the pre-trained language models. (AE: Auto-Encoding, AR: Auto-Regressive, T: Token, S: Segment, P: Position, W: Word, E: Entity, Ph: Phrase, WWM: Whole Word Masking, NM: N-gram Masking, NSP: Next Sentence Prediction, SOP: Sentence Order Prediction, MLM: Masked LM, PLM: Permutation LM, Mac: MLM as correction)}
\end{table*}

BERT \citep{devlin-etal-2019-bert} has proven to be successful in various NLU tasks, which is a representative auto-encoding PLM.
There are two pre-training tasks for BERT, shown as follows.
\begin{itemize}
	\item {\bf Masked Language Model (MLM)}: The MLM task randomly masks several tokens in the input sequence and requires to recover these tokens in the output. To correctly predict the original tokens, the model should utilize bidirectional context around the masked position.
	\item {\bf Next Sentence Prediction (NSP)}: The NSP task mainly focuses on the relationship in a larger context. It discriminates whether a sentence is the next sentence of another one.
\end{itemize}

To further improve the difficulties in the MLM task, \citet{devlin-etal-2019-bert} further proposed a technique called whole word masking (wwm). 
In this setting, instead of randomly selecting WordPiece \citep{wu2016google} tokens to mask, we always mask all of the tokens corresponding to a whole word at once. 
The whole word masking alleviates the ``input information leaking'' issue and has proven to be more effective than original MLM in various tasks \citep{chinese-bert-wwm}.
Furthermore, we can mask a consecutive N-gram to make MLM more difficult.
The N-gram masking has also proven to be effective in various PLMs, such SpanBERT \citep{joshi2019spanbert}, MacBERT \citep{cui-etal-2020-revisiting}, etc.

While MLM and its variants have been dominant in the design of various PLMs, it is intriguing to investigate other pre-training approaches other than MLM.
ELECTRA \citep{clark2020electra} employs a new generator-discriminator framework that is similar to GAN \citep{goodfellow-gan-nips2014}.
ELECTRA is largely different from BERT variants, but it still utilizes MLM in training the generator. 

In this paper, we take a step further on designing pre-training tasks. 
We design a pre-training task that does not adopt the MLM task, called permuted language model (PerLM). 
PerLM utilizes shuffled input text, and the objective is to predict the position of the original token. 
A detailed illustration of PERT is shown in the next section.

\section{PERT}

\subsection{Overview}
\begin{figure}[htp]
  \centering
  \includegraphics[width=1\columnwidth]{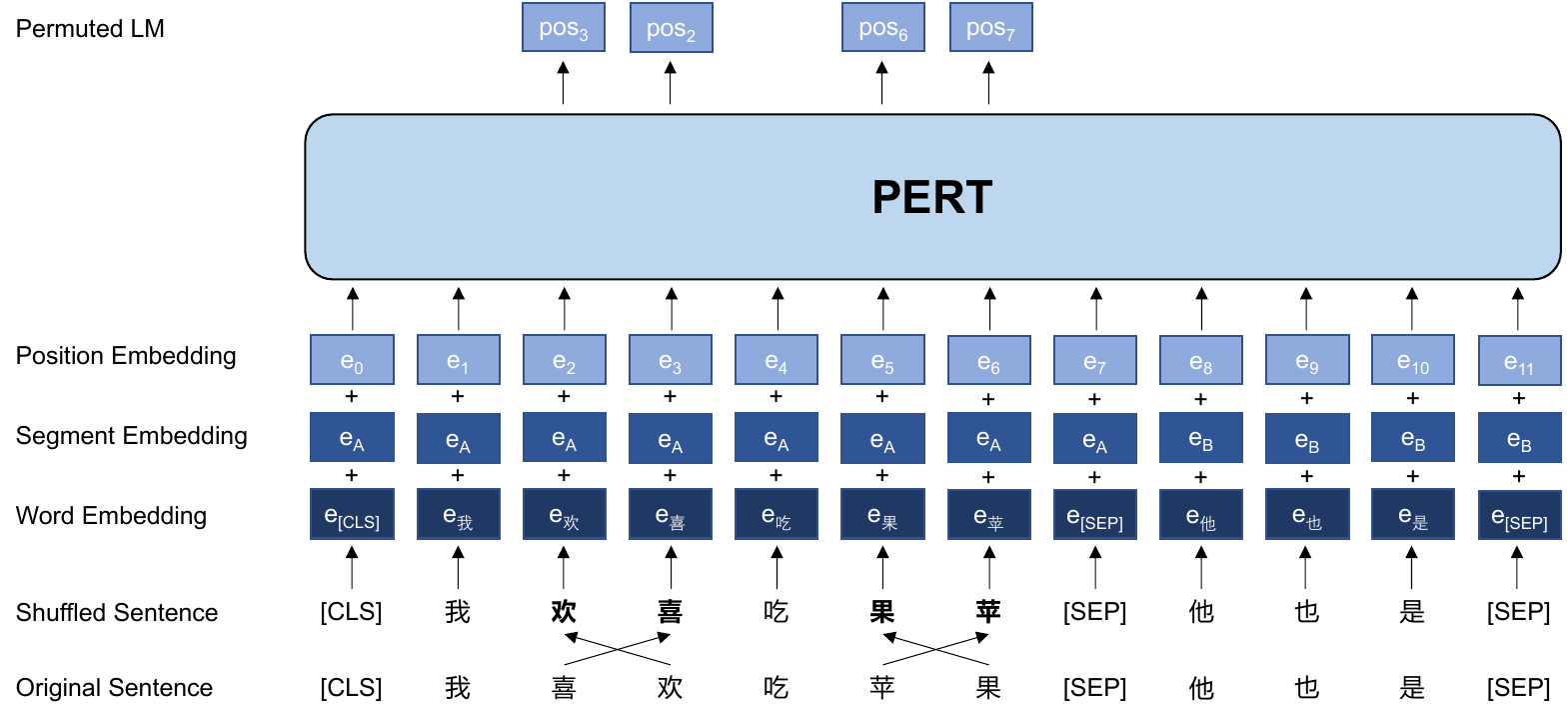}
  \caption{\label{pert-model} Neural architecture of PERT. } 
\end{figure}

An overview of PERT is depicted in Figure \ref{pert-model}.
As we can see that the proposed PERT shares identical neural architecture with BERT, while there is a slight difference in the input and the training objective. 
The proposed PERT uses a shuffled sentence\footnote{The term ``sentence'' here can be a consecutive text sequence and does not exactly represent the meaning in linguistic terminology.} as the input, and the training objective is to predict the position of the original token.
The main features of PERT are listed as follows.
\begin{itemize}
	\item Besides the replacement of MLM to PerLM, PERT largely adopts the original architecture of BERT, such as tokenization (using WordPiece), vocabulary (direct adoption), etc.
	\item PERT does not employ the artificial masking token {\tt [MASK]}.
	\item We use both whole word masking and N-gram masking to further enhance the performance.
	\item The prediction space is based on input sequence length but not the whole vocabulary (like MLM).
	\item Following previous works, we do not employ NSP-like pre-training tasks.
	\item As the main body of PERT is the same with BERT, BERT can be directly replaced by PERT with proper fine-tuning.
\end{itemize}

\subsection{Permuted Language Model}\label{sec-perlm}
As stated before, following various well-known PLMs, such as RoBERTa \citep{liu2019roberta}, we do not employ NSP-like pre-training tasks in PERT.
The only pre-training task used in our PERT is the proposed Permuted Language Model (PerLM).\footnote{To distinguish from the pre-training task (permutation LM) in XLNet \citep{yang2019xlnet}, we use the term ``permuted language model (PerLM)'' for our PERT in this paper.}
The formulation of PerLM is as follows.
\begin{itemize}
	\item We use whole word masking as well as N-gram masking strategies for selecting candidate tokens for masking, with a percentage of 40\%, 30\%, 20\%, 10\% for word-level unigram to 4-gram. This is identical to MacBERT \citep{chinese-bert-wwm} setting.
	\item Following previous works, we use a percentage of 15\% input words for masking purposes. Among which, 
	\begin{itemize}
		\item We randomly select a set of 90\% tokens and shuffle their orders. Note that the shuffle process only takes place for these 90\% tokens, not the whole input sequence.
		\item For the rest of 10\% tokens, we keep them unchanged, treating them as negative samples.
	\end{itemize}
\end{itemize}

As we can see that PerLM is as simple as the original MLM, while PerLM features the following characters.
\begin{itemize}
	\item PerLM does not employ the artificial token {\tt [MASK]} for masking purposes, which alleviates the pretraining-finetuning discrepancy issue (but could still suffer from unnatural word orders).
	\item The prediction space for PerLM is the input sequence, rather than on the whole vocabulary, making it computationally efficient than MLM.
\end{itemize}

\subsection{Pre-training Stage}
Formally, given a pair of sequences $A=\{A_1,\dots,A_n\}$ and $B=\{B_1,\dots,B_m\}$, we first use the method described in Section \ref{sec-perlm} to create new input sequence pairs $A'=\{A'_1,\dots,A'_n\}$ and $B'=\{B'_1,\dots,B'_m\}$, where some of the word positions are switched.
Then we concatenate two sequences to form the input sequence $X$ of PERT.
\begin{equation} 
X = {\tt [CLS]} ~{A'_1 \dots A'_n}~ {\tt [SEP]} ~{B'_1 \dots B'_m}~ {\tt [SEP]}
\end{equation}

Then, PERT converts $X$ into a contextualized representation $\bm{H} \in \mathbb{R}^{N  \times d}$ through an embedding layer, consisting of word embedding, positional embedding, and token type (segment) embedding, and a consecutive $L$-layer transformer, where $N$ is the maximum sequence length, and $d$ is the dimension of hidden layers.
\begin{gather} 
\bm{H}^{(0)} = \mathbf{Embedding}( X ) \\
\bm{H}^{(i)} = \mathbf{Transformer}(\bm{H}^{(i-1)}),~~i \in \{1,\dots,L\}, ~~ \bm{H} = \bm{H}^{(L)}
\end{gather}

Similar to MLM and Mac (MacBERT objective), we only need to predict the chosen positions in PerLM.
We gather a subset with respect to these positions, forming the candidate representation $\bm{H}^\text{m} \in \mathbb{R}^{k \times d}$, where $k$ is the number of the chosen tokens.
According to the definition of PerLM, we adopt a masking ratio of 15\%, and thus $k=\lfloor N \times 15\% \rfloor$.

Then we use a feed-forward dense layer (FFN), followed by a dropout and layer normalization layer.
\begin{gather} 
\tilde{\bm{H}}^{m} = \mathbf{LayerNorm}(\mathbf{Dropout}(\mathbf{FFN}(\bm{H}^{m}) )))
\end{gather}

To calculate the positions of the original tokens, we simply make a dot product between the $\tilde{\bm{H}}^{m}$ and $\bm{H}$.
Then we add a bias term $\bm{b} \in \mathbb{R}^{L}$ and use the softmax function to get normalized probabilities.
\begin{equation}
	\bm{p}_i = \mathbf{softmax}( \tilde{\bm{H}}^\text{m}_i \bm{H}^\top + \bm{b} ), ~~\bm{p}_i \in \mathbb{R}^{L}
\end{equation}

Finally, we use the standard cross-entropy loss to optimize the pre-training task.
\begin{equation}\label{equation-ce-loss}
	\mathcal{L} = -\frac{1}{M}\sum_{i=1}^M \bm{y}_i \log \bm{p}_i
\end{equation}

\subsection{Fine-tuning Stage}
PERT follows the same paradigm as in BERT to perform fine-tuning on various downstream tasks, as they share the same main neural architecture.
That is to say, PERT can directly fit in any fine-tuning script that is used for BERT or similar.
{\bf It is worth noting that, unlike the pre-training stage, we use natural input sequence rather than changing the word orders in the fine-tuning stage.}

\section{Experiments on Chinese Tasks}

\subsection{Pre-training Setups}
We largely follow the training recipe of MacBERT, where we illustrate as follows. All models are trained from scratch.
\begin{itemize}
	\item {\bf Data}: We use the training data as in MacBERT. It consists of the Chinese Wikipedia dump\footnote{https://dumps.wikimedia.org/zhwiki/latest/}, encyclopedia, community question answering, news articles, etc. The total training data has 5.4B words and takes about 20G disk space.
	\item {\bf Tokenization}: We use WordPiece tokenizer \citep{wu2016google} as in BERT. To detect the Chinese word boundaries, we use LTP \citep{che2010ltp} for word segmentation. Note that the Chinese word segmentation is only used for selecting the whole word to perform whole word masking, i.e., only affect which tokens are chosen for masking. The input for PERT is still handled by the WordPiece tokenizer.
	\item {\bf Vocabulary}: We directly use the vocabulary of Chinese BERT-base\footnote{https://storage.googleapis.com/bert\_models/2018\_11\_03/chinese\_L-12\_H-768\_A-12.zip} and other PLMs with a vocabulary size of 21128.
	\item {\bf Hyper-parameters}: We use a maximum sequence length of 512 throughout the whole pre-training process. 
	\item {\bf Optimization}: We use a batch size of 416 (base-level) or 128 (large-level) with an initial learning rate of 1e-4. We perform a linear warmup schedule with the first 10K steps. The total training step is 2M. We use \textsc{Adam} \citep{kingma2014adam} with weight decay (rate = 0.1) optimizer with beta values (0.9, 0.999) and an epsilon value 1e-6. 
	\item {\bf Training Device}: The training was done on a single TPU v3-8 (128G HBM).\footnote{https://cloud.google.com/tpu/}
\end{itemize}

Following previous works, we train two PERT models: PERT-base (12-layer, 12-heads, 768-dim) and PERT-large (24-layer, 16-heads, 1024-dim), which are the same with BERT settings.

\subsection{Fine-tuning Setups}
We choose the following ten popular Chinese NLU datasets.
The fine-tuning settings and statistics for each task are shown in Table \ref{chinese-tasks}.
\begin{itemize}[leftmargin=*]
	\item {\bf Machine Reading Comprehension (MRC)}: CMRC 2018 \citep{cui-emnlp2019-cmrc2018}, DRCD \citep{shao2018drcd}.
	\item {\bf Text Classification (TC)}: XNLI \citep{conneau2018xnli}, LCQMC \citep{liu2018lcqmc}, BQ Corpus \citep{chen-etal-2018-bq}, ChnSentiCorp \citep{tan2008empirical}, TNEWS \citep{clue}, OCNLI \citep{ocnli}. 
	\item {\bf Named Entity Recognition (NER)}: MSRA-NER (SIGHAN 2006) \citep{levow-2006-third}, People's Daily\footnote{https://github.com/ProHiryu/bert-chinese-ner/tree/master/data}.
\end{itemize}

\begin{table*}[h]
\small
\begin{center}
\begin{tabular}{c | l | c c c c | c c c}
\toprule
\bf Task & \bf Dataset & \bf MaxLen & \bf Batch & \bf Epoch & \bf InitLR & \bf Train \# & \bf Dev \# & \bf Test \#  \\
\midrule
\multirow{2}*{\bf MRC} & CMRC 2018 	 & 512 & 64 & 2 & 3e-5 & 10K & 3.2K & 4.9K  \\
& DRCD 		 & 512 & 64 & 2 & 3e-5 & 27K & 3.5K & 3.5K  \\
\midrule
\multirow{6}*{\bf TC} & XNLI& 128 & 64 & 2 & 3e-5 & 392K & 2.5K & 5K  \\
& LCQMC 		 & 128 & 64 & 3 & 2e-5 & 240K & 8.8K & 12.5K  \\
& BQ Corpus	  	 & 128 & 64 & 3 & 3e-5 & 100K & 10K & 10K  \\
& ChnSentiCorp 	 & 256 & 64 & 3 & 2e-5 & 9.6K & 1.2K & 1.2K \\
& TNEWS			 & 128 & 64 & 3 & 2e-5 & 53.3K & 10K & 10K \\
& OCNLI 		 & 128 & 64 & 3 & 2e-5 & 56K & 3K & 3K \\
\midrule
\multirow{2}*{\bf NER} 	& MSRA-NER  & 256 & 64 & 5 & 3e-5 & 45K & - & 3.4K \\
& People's Daily	 					& 256 & 64 & 5 & 3e-5 & 51K & 4.6K & -\\
\bottomrule
\end{tabular}
\end{center}
\caption{\label{chinese-tasks} Hyper-parameter settings and data statistics for Chinese tasks.}
\end{table*}

For fair comparisons, we use the models trained on the same pre-training corpora (20G). 
These models include: BERT$_\text{base}$ (i.e., BERT-wwm-ext\footnote{https://github.com/ymcui/Chinese-BERT-wwm}), RoBERTa$_\text{base}$ (i.e., RoBERTa-wwm-ext\footnote{https://github.com/ymcui/Chinese-BERT-wwm}), ELECTRA$_\text{base}$\footnote{https://github.com/ymcui/Chinese-ELECTRA}, MacBERT$_\text{base}$\footnote{https://github.com/ymcui/MacBERT}.
Also, to ensure robust experimentation, we carry out each experiment ten times with different random seeds and report both the maximum and average scores for all tasks, except for the grammar checking tasks.
The fine-tuning scripts are based on original BERT implementation, including {\tt run\_classifier.py} for classification tasks, and {\tt run\_squad.py} for MRC tasks.\footnote{https://github.com/google-research/bert}

\subsection{Results on Machine Reading Comprehension Tasks}
The results on machine reading comprehension (MRC) tasks are shown in Table \ref{result-chinese-mrc}.
The evaluation metrics for MRC tasks are the exact match (EM) and F1. 
As we can see that the proposed PERT yields moderate improvements over MacBERT and is consistently outperform the others, setting state-of-the-art performances on several subsets.
This demonstrates that by permuting words in the input sequence, PERT learns both short-range and long-range text inference abilities, which is critical in machine reading comprehension tasks.
\begin{table*}[h]
\tiny
\begin{center}
\begin{tabular}{l c c c c c c | c c c c}
\toprule
\multirow{2}*{\bf System} & \multicolumn{6}{c}{\centering \bf CMRC 2018} & \multicolumn{4}{c}{\centering \bf DRCD} \\
& \bf D-EM & \bf D-F1 & \bf T-EM & \bf T-F1  & \bf C-EM & \bf C-F1  & \bf D-EM & \bf D-F1  & \bf T-EM & \bf T-F1  \\
\midrule
BERT$_\text{base}$ & 67.1 \tiny(65.6) & 85.7 \tiny(85.0) & 71.4 \tiny(70.0) & 87.7 \tiny(87.0) & 24.0 \tiny(20.0) & 47.3 \tiny(44.6) & 85.0 \tiny(84.5) & 91.2 \tiny(90.9) & 83.6 \tiny(83.0) & 90.4 \tiny(89.9)  \\ 
RoBERTa$_\text{base}$ & 67.4 \tiny(66.5) & 87.2 \tiny(86.5) & 72.6 \tiny(71.4) & 89.4 \tiny(88.8) & 26.2 \tiny(24.6) & 51.0 \tiny(49.1) & 86.6 \tiny(85.9) & 92.5 \tiny(92.2) & 85.6 \tiny(85.2) & 92.0 \tiny(91.7)  \\
ELECTRA$_\text{base}$ & 68.4 \tiny(68.0) & 84.8 \tiny(84.6) & 73.1 \bf\tiny(72.7) & 87.1 \tiny(86.9) & 22.6 \tiny(21.7) & 45.0 \tiny(43.8) & 87.5 \tiny(87.0) & 92.5 \tiny(92.3) & 86.9 \tiny(86.6) & 91.8 \tiny(91.7) \\
MacBERT$_\text{base}$ & {\bf 68.5} \tiny(67.3) & \bf 87.9 \tiny(87.1) & {\bf 73.2} \tiny(72.4) & \bf 89.5 \tiny(89.2) & {\bf 30.2} \tiny(26.4) & 54.0 \tiny(52.2) & 89.4 \bf \tiny(89.2) & \bf 94.3 \tiny(94.1) & \bf 89.5 \tiny(88.7) & \bf 93.8 \tiny(93.5) \\
\bf PERT$_\text{base}$ & \bf 68.5 \tiny(68.1) & 87.2 \bf \tiny(87.1) & 72.8 \tiny(72.5) & 89.2 \tiny(89.0) & 28.7 \bf \tiny(28.2) & \bf 55.4 \tiny(53.7) & {\bf 89.5} \tiny(88.9) & 93.9 \tiny(93.6) & 89.0 (88.5) & 93.5 \tiny(93.2) \\
\midrule 
RoBERTa$_\text{large}$ & 68.5 \tiny(67.6) & 88.4 \tiny(87.9) & 74.2 \tiny(72.4) & 90.6 \tiny(90.0) & 31.5 \tiny(30.1) & 60.1 \tiny(57.5)  & 89.6 \tiny(89.1) & 94.8 \tiny(94.4) & 89.6 \tiny(88.9) & 94.5 \tiny(94.1) \\
ELECTRA$_\text{large}$ & 69.1 \tiny(68.2) & 85.2 \tiny(84.5) & 73.9 \tiny(72.8) & 87.1 \tiny(86.6) & 23.0 \tiny(21.6) & 44.2 \tiny(43.2) & 88.8 \tiny(88.7) & 93.3 \tiny(93.2) & 88.8 \tiny(88.2) & 93.6 \tiny(93.2) \\
MacBERT$_\text{large}$ 		& 70.7 \tiny(68.6) & 88.9 \tiny(88.2) & 74.8 \tiny(73.2) & {\bf 90.7} \tiny(90.1) & 31.9 \tiny(29.6) & {\bf 60.2} \tiny(57.6) & \bf 91.2 \tiny(90.8) & \bf 95.6 \tiny(95.3) & \bf 91.7 \tiny(90.9) & \bf 95.6 \tiny(95.3) \\
\bf PERT$_\text{large}$ & \bf 72.2 \tiny(71.0) & \bf 89.4 \tiny(88.8) & \bf 76.8 \tiny(75.5) & \bf 90.7 \tiny(90.4) & \bf 32.3 \tiny(30.9) & 59.2 \bf \tiny(58.1) & 90.9 \bf \tiny(90.8) & 95.5 \tiny(95.2) & 91.1 \tiny(90.7) & 95.2 \tiny(95.1) \\
\bottomrule
\end{tabular}
\end{center}
\caption{\label{result-chinese-mrc} Results on Chinese machine reading comprehension (MRC) tasks: CMRC 2018 (Simplified Chinese) and DRCD (Traditional Chinese). We report the maximum and average scores (in brackets) for each set. Overall best performances are depicted in boldface (base-level and large-level are marked individually). D: Dev set, T: Test set, C: Challenge set.}
\end{table*}

\subsection{Results on Text Classification Tasks}
The results on text classification (TC) tasks are shown in Table \ref{result-chinese-tc}.
Unfortunately, the proposed PERT does not perform well on text classification tasks. 
We conjecture that the permuted input text in the pre-training stage brings difficulties in understand short text compared to MRC tasks. 
\begin{table*}[h]
\tiny
\begin{center}
\begin{tabular}{l cc cc cc cc cc }
\toprule
\multirow{2}*{\bf System} & \multicolumn{2}{c}{\centering \bf XNLI} & \multicolumn{2}{c}{\centering \bf LCQMC} & \multicolumn{2}{c}{\centering \bf BQ Corpus} & \multicolumn{2}{c}{\centering \bf ChnSentiCorp}  & \bf TNEWS  & \bf OCNLI \\
 & \bf Dev & \bf Test  & \bf Dev & \bf Test & \bf Dev & \bf Test & \bf Dev & \bf Test & \bf Dev & \bf Dev \\
\midrule
BERT$_\text{base}$ & 79.4 \tiny(78.6) & 78.7 \tiny(78.3)  & 89.6 \tiny(89.2) & 87.1 \tiny(86.6) & {\bf 86.4} \tiny(85.5)  & {\bf 85.3} \tiny(84.8) & {\bf 95.4} \tiny(94.6) & 95.3 \tiny(94.8)  & 57.0 \tiny(56.6) & 76.0 \tiny(75.3) \\
RoBERTa$_\text{base}$ & 80.0 \tiny(79.2) & 78.8 \tiny(78.3)  & 89.0 \tiny(88.7) & 86.4 \tiny(86.1) & 86.0 \tiny(85.4) & 85.0 \tiny(84.6) & 94.9 \tiny(94.6) & {\bf 95.6} \tiny(94.9)   & {\bf 57.4} \tiny(56.9)  & 76.5 \tiny(76.0)  \\
ELECTRA$_\text{base}$ 	& 77.9 \tiny(77.0) & 78.4 \tiny(77.8) & \bf 90.2 \tiny(89.8) & \bf 87.6 \tiny(87.3) & 84.8 \tiny(84.7) & 84.5 \tiny(84.0) & 93.8 \tiny(93.0) & 94.5 \tiny(93.5)  & 56.1 \tiny(55.7)  & 76.1 \tiny(75.8) \\  
MacBERT$_\text{base}$ & \bf 80.3 \tiny(79.7) & \bf 79.3 \tiny(78.8) & 89.5 \tiny(89.3) & 87.0 \tiny(86.5) & 86.0 \tiny(85.5) & 85.2 \bf\tiny(84.9) & 95.2 \bf\tiny(94.8) & {\bf 95.6} \tiny(94.9)  & \bf 57.4 \tiny(57.1)  & \bf 77.0 \tiny(76.5)  \\
\bf PERT$_\text{base}$ & 78.8 \tiny(78.1) & 78.1 \tiny(77.7) & 88.8 \tiny(88.3) & 86.3 \tiny(86.0) & 84.9 \tiny(84.8) & 84.3 \tiny(84.1) & 94.0 \tiny(93.7) & 94.8 \tiny(94.1) & 56.7 \tiny(56.1) & 75.3 \tiny(74.8) \\
\midrule
RoBERTa$_\text{large}$ & 82.1 \tiny(81.3) & 81.2 \tiny(80.6)  & 90.4 \tiny(90.0) & 87.0 \tiny(86.8) & 86.3 \tiny(85.7) & {\bf 85.8} \tiny(84.9) & {\bf 95.8} \tiny(94.9) & 95.8 \tiny(94.9)  & 58.8 \tiny(58.4)  & 78.5 \tiny(78.2)  \\
ELECTRA$_\text{large}$ 	& 81.5 \tiny(80.8) & 81.0 \bf\tiny(80.9) & \bf 90.7 \tiny(90.4) & 87.3 \bf\tiny(87.2) & \bf 86.7 \tiny(86.2) & 85.1 \tiny(84.8) & 95.2 \tiny(94.6) & 95.3 \tiny(94.8)   & 57.2 \tiny(56.9) & 78.8 \tiny(78.4)  \\
MacBERT$_\text{large}$ & \bf 82.4 \tiny(81.8) & {\bf 81.3} \tiny(80.6) & 90.6 \tiny(90.3) & {\bf 87.6} \tiny(87.1) & 86.2 \tiny(85.7) & 85.6 \bf \tiny(85.0) & 95.7 \bf \tiny(95.0) & {\bf 95.9} \bf\tiny(95.1)  & \bf 59.0 \tiny(58.8) & \bf 79.0 \tiny(78.7) \\
\bf PERT$_\text{large}$ & 81.0 \tiny(80.4) & 80.4 \tiny(80.1) & 90.0 \tiny(89.7) & 87.2 \tiny(86.9) & 86.3 \tiny(85.8) & 85.0 \tiny(84.8) & 94.5 \tiny(94.0) & 95.3 \tiny(94.8) & 57.4 \tiny(57.2) & 78.1 \tiny(77.8) \\
\bottomrule
\end{tabular}
\end{center}
\caption{\label{result-chinese-tc} Results on text classification (TC) tasks: XNLI, LCQMC, BQ Corpus, ChnSentiCorp, TNEWS, and OCNLI. }
\end{table*}

\subsection{Results on Named Entity Recognition Tasks}
The results on named entity recognition (NER) tasks are shown in Table \ref{result-chinese-ner}.
We extract the predicted entities and use {\tt seqeval}\footnote{https://github.com/chakki-works/seqeval} to evaluate the NER performance in terms of P/R/F metrics.
As we can see that PERT yields relatively consistent improvements over all baseline systems, indicating its good abilities in sequence tagging tasks.

Based on all the experiments above, we make several conclusions as follows. 
1) PERT yields better performances on MRC and NER tasks, but it does not perform well on TC tasks; 
2) The PerLM with whole word masking and N-gram masking make PERT more sensitive to the word/phrase boundaries, which is helpful in span-extraction MRC and NER tasks;
3) The input sequence for pre-training PERT is shuffled to some extent. According to the results of TC tasks, PERT yields inferior performances. This means text classification tasks are more sensitive to the word permutation. 

Some of the word permutation will bring a complete meaning change for the input text.
This will affect all the fine-tuning tasks presented above. 
However, TC tasks suffer from this issue more than MRC or NER tasks, as the input text of TC tasks is relatively shorter than the others.
MRC and NER tasks also suffer from word permutation. 
However, the input text for MRC tasks is typically long, and several word permutations may not change the whole narrative flows of the passage. 
For NER tasks, such permutation may not affect the NER process, as the named entities only take a small proportion in the whole input text.

\begin{table*}[h]
\small
\begin{center}
\begin{tabular}{l  ccc ccc }
\toprule
\multirow{2}*{\bf System} & \multicolumn{3}{c}{\centering \bf MSRA-NER (Test)} & \multicolumn{3}{c}{\centering \bf People's Daily (Dev)}  \\
& \bf P  & \bf R & \bf F & \bf P & \bf R & \bf F \\
\midrule
BERT$_\text{base}$ 		& 95.2 \tiny(94.8) & 95.4 \tiny(95.1) & 95.3 \tiny(94.9) & 95.3 \bf \tiny (95.1) & 95.3 \tiny (95.1) & \bf 95.3 \tiny (95.1) \\
RoBERTa$_\text{base}$ 	& 95.3 \tiny(94.9) & 95.6 \tiny(95.4) & 95.5 \tiny(95.1) & 94.9 \tiny(94.8) & 95.3 \tiny(95.1) & 95.1 \tiny(94.9) \\
ELECTRA$_\text{base}$ 	& 95.0 \tiny(94.5) & {\bf 95.9} \tiny(95.4) & 95.4 \tiny(95.0) & 94.8 \tiny(94.7) & 95.3 \bf\tiny(95.2) & 95.1 \tiny(94.9) \\  
MacBERT$_\text{base}$ 	& 95.2 \tiny(94.9) & 95.4 \tiny(95.4) & 95.3 \tiny(95.1) & 94.9 \tiny(94.6) & 95.6 \tiny(95.1) & 95.2 \tiny(94.9) \\
\bf PERT$_\text{base}$ 	& \bf 95.4 \tiny(95.2) & 95.5 \bf \tiny(95.5) & \bf {\bf 95.6} \tiny(95.3) & \bf 95.4 \tiny(95.1) & 95.2 \tiny(95.0) & \bf 95.3 \tiny(95.1) \\
\midrule
RoBERTa$_\text{large}$ 	& 95.4 \tiny(95.3) & 95.7 \tiny(95.7) & 95.5 \tiny(95.5) & 95.7 \tiny(95.4) & 95.7 \tiny(95.4) & 95.7 \tiny(95.4) \\
ELECTRA$_\text{large}$ 	& 94.9 \tiny(94.8) & 95.5 \tiny(95.0) & 95.0 \tiny(94.8) & 94.8 \tiny(94.6) & 95.3 \tiny(95.3) & 94.9 \tiny(94.8) \\  
MacBERT$_\text{large}$ 	& 96.3 \tiny(95.8) & 96.3 \tiny(95.9) & {\bf 96.2} \tiny(95.9) & 95.8 \tiny(95.6) & 95.8 \bf \tiny(95.7)  & 95.8 \tiny(95.7) \\
\bf PERT$_\text{large}$ & \bf 96.4 \tiny(95.9) & \bf 96.4 \tiny(96.1) & \bf 96.2 \tiny(96.0) & \bf  96.3 \bf \tiny(96.0) & 96.0 \bf \tiny(95.7) & \bf 96.1 \tiny(95.8) \\
\bottomrule
\end{tabular}
\end{center}
\caption{\label{result-chinese-ner} Results on Chinese named entity recognition (NER) tasks.}
\end{table*}

\subsection{Results on Grammar Checking Tasks}
Besides traditional public Chinese NLU tasks, we also test PERT on in-house grammar checking tasks. 
Specifically, we focus on {\em Word Order Recovery} (WOR) task, which is a part of the grammar checking system\footnote{http://check.hfl-rc.com}.

\subsubsection{Task Definition}
The objective of the WOR task is to fix the grammar errors caused by incorrect word orders.
Specifically, in the WOR task, the examples are the sentences where some words have been moved to incorrect positions.
For example, in the sentence ``我每天一个吃苹果 (I everyday an eat apple)'' there is an error span ``一个 (an)吃 (eat)'', where the order of ``一个(an)'' and ``吃(eat)'' have been swapped. WOR task asks the model to recover the correct sentence ``我每天吃一个苹果 (I everyday eat an apple).''
A sentence may contain multiple error spans. 
To simplify the problem, we only consider the case where only one word has been moved in each span.

\subsubsection{Modeling}
We treat the WOR task as a sequence labeling task and take the ``BIEO'' tagging scheme. 
The inputs to the model are incorrect sentences; the label of each word in the sentences stands for whether the word is in the correct position (O), or it is the beginning of an error span (B), or it should be moved to the last of the error span (I), or it should be moved to the beginning of the error span (E). 
For example, the sentence ``我每天一个吃苹果'' is labeled as ``我(O)每(O)天(O)一(B)个(I)吃(E)苹(O)果(O)''.
With the tagging scheme above, we can easily recover the correct word orders in the sentence.

\subsubsection{Results}
We mainly test WOR task under four domains (train/dev): Wikipedia (990K/86K), Formal Doc. (1.4M/33K), Customs (682K/34K), Legal (1.8M/13K).
We report precision, recall, and F1 scores for the following experiments.
The results are shown in Table \ref{result-chinese-grammar}.
PERT yields consistent and significant improvements over all baseline systems in terms of all evaluation metrics (P/R/F). 
This is in accordance with our expectations, as the fine-tuning task is quite similar to the pre-training task of PERT. 
Though the fine-tuning is performed in a sequence tagging manner (just like NER), it still can benefit from the pre-training of PERT, which focuses on ordering the words in the correct position.
\begin{table*}[h]
\small
\begin{center}
\begin{tabular}{l ccc | ccc | ccc | ccc }
\toprule
\multirow{2}*{\bf System} & \multicolumn{3}{c}{\centering \bf Wikipedia} & \multicolumn{3}{c}{\centering \bf Formal Doc.} & \multicolumn{3}{c}{\centering \bf Customs} & \multicolumn{3}{c}{\centering \bf Legal}  \\
& \bf P  & \bf R & \bf F & \bf P & \bf R & \bf F & \bf P & \bf R & \bf F & \bf P & \bf R & \bf F \\
\midrule
BERT$_\text{base}^\text{Google}$ & 83.6 & 76.3 & 79.8 & 92.1 & 87.1 & 89.6 & 85.7 & 85.1 & 85.4 & 94.3 & 89.8 & 92.0 \\
RoBERTa$_\text{base}$ 		& 84.2 & 76.9 & 80.4 & 92.6 & 87.7 & 90.1 & 86.8 & 85.9 & 86.3 & 94.6 & 90.0 & 92.2 \\
ELECTRA$_\text{base}$ 		& 69.9 & 57.8 & 63.6 & 88.1 & 81.6 & 84.7 & 69.6 & 71.0 & 70.3 & 91.7 & 85.4 & 88.4 \\
MacBERT$_\text{base}$ 		& 84.3 & 77.1 & 80.5 & 92.7 & 87.8 & 90.2 & 86.4 & 86.5 & 86.4 & 94.6 & 90.1 & 92.3 \\
\bf PERT$_\text{base}$ 		& \bf 86.5 & \bf 79.5 & \bf 82.9 & \bf 93.6 & \bf 89.0 & \bf 91.2 & \bf 88.3 & \bf 88.0 & \bf 88.2 & \bf 95.2 & \bf 90.7 & \bf 92.9 \\
\bottomrule
\end{tabular}
\end{center}
\caption{\label{result-chinese-grammar} Results on in-house grammar checking tasks.}
\end{table*}

\section{Experiments on English Tasks}

\subsection{Pre-training Setups}
We largely follow the training recipe of BERT, where we illustrate as follows. All models are trained from scratch.
\begin{itemize}
	\item {\bf Data}: We use English Wikipedia and BooksCorpus \citep{zhu2015aligning} as the pre-training data, which is widely used in the previous literature. 
	\item {\bf Tokenization}: We use WordPiece tokenizer \citep{wu2016google} as in BERT. 
	\item {\bf Vocabulary}: We directly use the vocabulary of English BERT-base-uncased\footnote{https://storage.googleapis.com/bert\_models/2018\_10\_18/uncased\_L-12\_H-768\_A-12.zip} with a vocabulary size of 30522.
	\item {\bf Hyper-parameters}: We use a maximum sequence length of 512 throughout the whole pre-training process. 
	\item {\bf Optimization}: We use a batch size of 416 (base-level) or 128 (large-level) with an initial learning rate of 1e-4. We perform a linear warmup schedule with the first 10k steps. The total training step is 2M. We use \textsc{Adam} \citep{kingma2014adam} with weight decay (rate = 0.1) optimizer with beta values (0.9, 0.999) and an epsilon value 1e-6. 
	\item {\bf Training Device}: The training was done on a single TPU v3-8 (128G HBM).
\end{itemize}

Similar to Chinese PERT, we also train two PERT models: PERT-base and PERT-large.

\subsection{Fine-tuning Setups}
We choose the following six popular English NLU datasets: SQuAD \citep{rajpurkar-etal-2016}, SQuAD 2.0 \citep{rajpurkar-etal-2018-know}, MNLI \citep{williams-etal-2018-broad}, SST-2 \citep{socher2013recursive}, CoLA \citep{warstadt2019neural}, and MRPC \citep{dolan-brockett-2005-automatically}. 
The fine-tuning settings for each task are shown in Table \ref{english-tasks}.
\begin{table*}[h]
\small
\begin{center}
\begin{tabular}{l c c c c | c c }
\toprule
\bf Dataset & \bf MaxLen & \bf Batch & \bf Epoch & \bf InitLR & \bf Train \# & \bf Dev \# \\
\midrule
SQuAD 1.1	& 512 & 64 & 2 & 3e-5 & 87.6K & 10.6K  \\
SQuAD 2.0	& 512 & 64 & 2 & 3e-5 & 130.3K & 11.9K \\
\midrule
MNLI 		& 256 & 64 & 3 & 3e-5 & 392.7K & 9.8K  \\
SST-2 		& 128 & 64 & 3 & 3e-5 & 67.3K & 0.9K  \\
CoLA	  	& 128 & 64 & 10 & 2e-5 & 8.6K & 1.0K  \\
MRPC 	 	& 512 & 64 & 5 & 3e-5 & 3.7K & 0.4K \\
\bottomrule
\end{tabular}
\end{center}
\caption{\label{english-tasks} Data statistics and hyper-parameter settings for English tasks.}
\end{table*}

For fair comparisons, we only list those model variants that were trained on WikiBooks (Wikipedia + BooksCorpus).
Note that, for all experiments on English tasks, we do not use additional tricks that were used in other papers, such as data augmentation, fine-tuning from MNLI checkpoints, etc.
We report five-run average scores for our experiments.

\subsection{Results}
We report EM/F1 for SQuAD and SQuAD 2.0, accuracy for MNLI, SST-2, and MRPC, Matthews correlation for CoLA.
The results are shown in Table \ref{results-english}.
Similar to the results of Chinese NLU tasks, PERT$_\text{base}$ yields better performance on MRC tasks and moderate improvements on a few TC tasks.
However, we noticed that PERT$_\text{large}$ performs worse than the others on most of the tasks. 
\begin{table*}[h]
\small
\begin{center}
\begin{tabular}{l  cccc | cccc }
\toprule
\multirow{2}*{\bf System} & \multicolumn{2}{c}{\centering \bf SQuAD} & \multicolumn{2}{c}{\centering \bf SQuAD 2.0} & \bf MNLI & \bf SST-2 & \bf CoLA & \bf MRPC \\
& \bf EM  & \bf F1 & \bf EM & \bf F1 & \bf Acc & \bf Acc & \bf M.C. & \bf Acc \\
\midrule
BERT$_\text{base}$			& 80.8 & 88.5 & - & - & 84.4 & \bf 92.7 & 60.6 & 86.7 \\
BERT$_\text{base}$$^\dag$ 	& 81.2 & 88.5 & 72.4 & 75.4 & 84.4 & 92.6 & 59.3 & 86.0 \\
RoBERTa$_\text{base}$		& - & 90.4 & - & 79.1 & 84.7 & 92.5 & - & - \\
XLNet$_\text{base}$			& - & - & \bf 78.4 & \bf 81.3 & \bf 85.8 & 92.6 & - & - \\
ALBERT$_\text{base}$			& 82.1 & 89.3 & 76.1 & 79.1 & 81.9 & 89.4 & - & - \\
\bf PERT$_\text{base}$		& \bf 84.8 & \bf 91.3 & 78.3 & 81.0 & 84.5 & 92.0 & \bf 61.2 & \bf 87.5 \\
\midrule
BERT$_\text{large}$				& 84.1 & 90.9 & 78.7 & 81.9 & 86.6 & 93.2 & 60.6 & 88.0 \\
BERT$_\text{large-wwm}$$^\dag$ 	& 87.4 & 93.4 & 82.8 & 85.6 & 87.3 & 93.4 & 63.1  & 87.2 \\
RoBERTa$_\text{large}$			& - & 93.6 & - & 87.3 & \bf 89.0 & \bf 95.3 & - & - \\
XLNet$_\text{large}$				& \bf 88.2 & \bf 94.0 & \bf 85.1 & \bf 87.8 & 88.4 & 94.4 & 65.2 & \bf 90.0 \\
ALBERT$_\text{large}$			& 84.1 & 90.9 & 79.0 & 82.1 & 83.8 & 90.6 & - & - \\
\bf PERT$_\text{large}$			& 87.4 & 93.3 & 83.5 & 86.3 & 87.6 & 93.4 & \bf 65.7 & 87.3 \\
\bottomrule
\end{tabular}
\end{center}
\caption{\label{results-english} Development set results on English NLU tasks. The system marked with $^\dag$ means the reproduced results (rerun). M.C.: Matthews correlation. }
\end{table*}

\section{Analysis}
In this section, we will look deeper into PERT with quantitative analyses.
We use Chinese PERT$_\text{base}$ for all analyses in the following subsections.

\subsection{Performance on Different Training Steps}
\begin{figure}[h]
  \centering
  \subfigure[MRC]{\includegraphics[width=0.32\columnwidth]{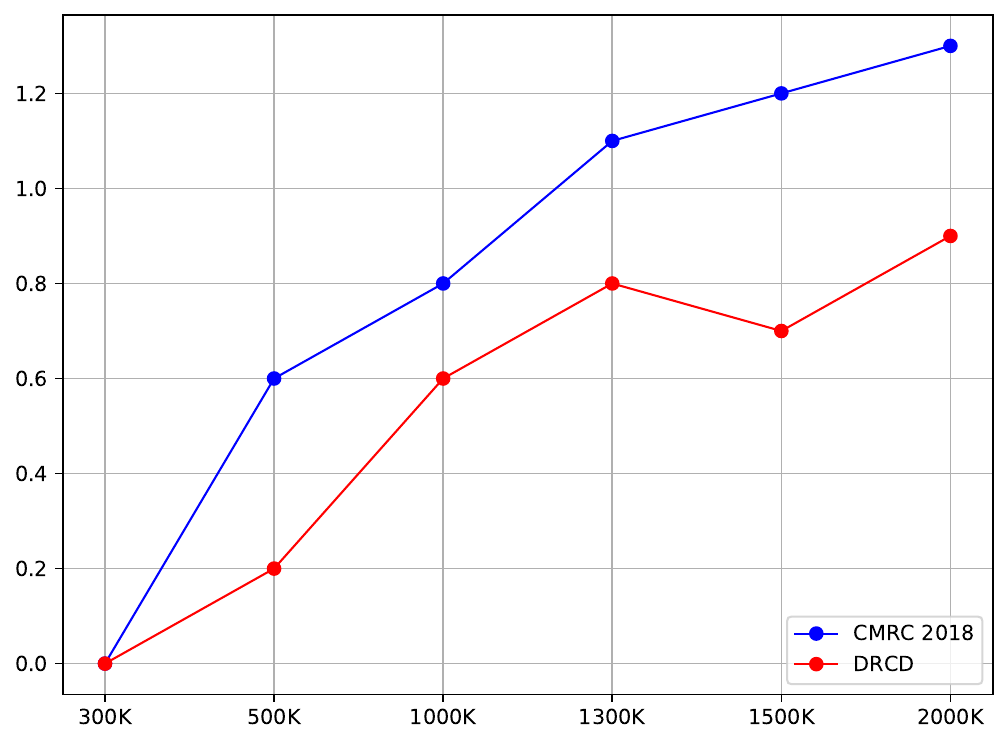}} 
  \subfigure[TC]{\includegraphics[width=0.32\columnwidth]{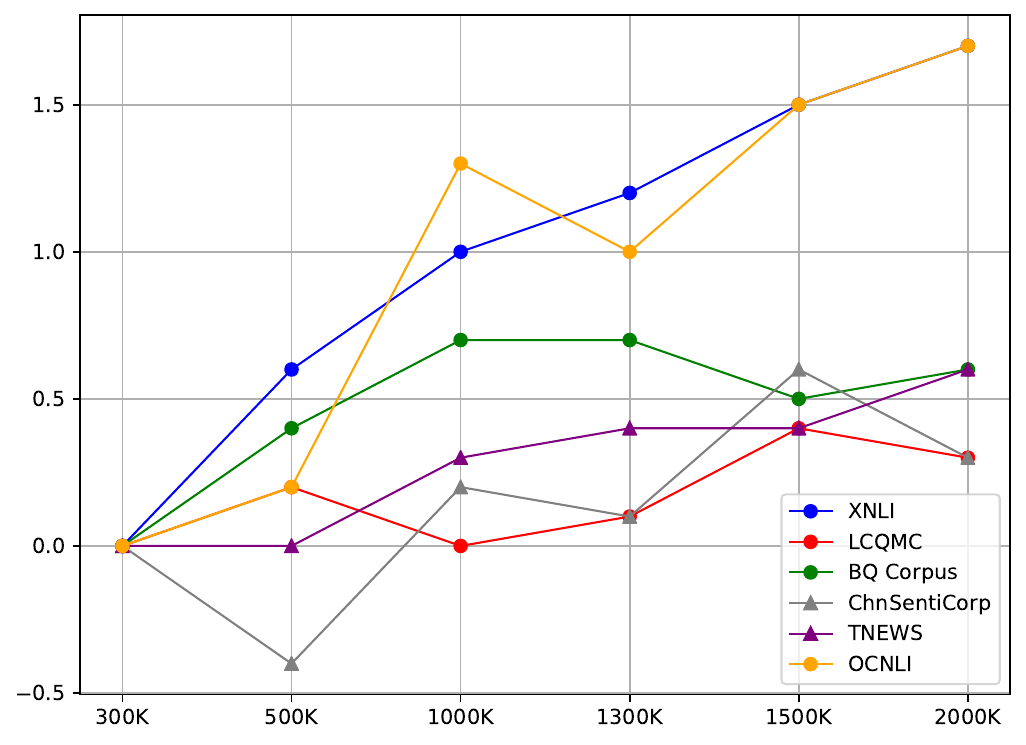}}
  \subfigure[NER]{\includegraphics[width=0.32\columnwidth]{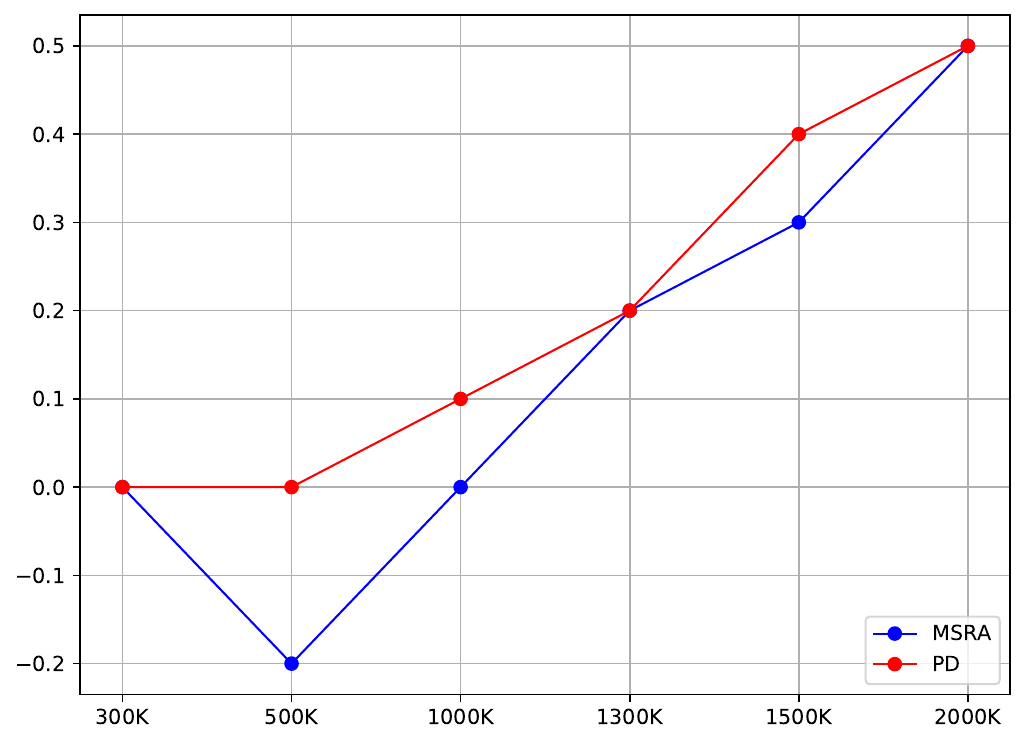}}
  \caption{\label{analysis-mrc-tc-ner} Performance on MRC, TC, and NER tasks on different pre-training steps. } 
\end{figure}

Just like the performance curves for fine-tuning tasks, the optimum performance for each fine-tuning task may not happen at the same pre-training step.
To investigate the performance of different types of the fine-tuning tasks, we plot their results on 300K, 500K, 1000K, 1300K, 1500K, and 2000K pre-training steps.
We use the performance of 300K as the baseline and calculate the gap to the baseline in terms of different training steps, where a positive value means a performance growth and negative values for decrease.
We report F1 for MRC tasks, accuracy for TC tasks, and F1 for NER tasks.
The results are shown in Figure \ref{analysis-mrc-tc-ner}.

For MRC and NER tasks, we see relatively consistent performance improvements with the growth of pre-training steps.
However, for TC tasks, the performance for some tasks reaches their best at 1000K or 1300K pre-training steps, such as BQ Corpus and TNEWS.
These results indicate that it is necessary to ``harvest'' the pre-trained model in an earlier pre-training step if the performance of a specific task is considered.

\subsection{Different Permutation Granularities}\label{sec-analysis-granularity}
In the formulation of PERT, we did not make restrictions on the permuting granularity, i.e., we directly permute input text in token-level, and any word can be replaced by another word in the passage (though the replacement is restricted {\em within} the selected tokens).
However, {\em what if we permute them within a complete word/N-gram/sentence?}
Such restrictions will make the input text more readable as the permutation is done within a linguistic unit rather than the whole sequence.
We compare the performance by permuting the input text within different granularities.
We pre-train PERT$_\text{base}$ with 300k steps and observe their performances on CMRC 2018, XNLI, TNEWS, and OCNLI tasks.
The results are shown in Table \ref{analysis-different-granularity}.
\begin{table*}[h]
\tiny
\begin{center}
\begin{tabular}{l c c c c c c | c c c c | c}
\toprule
\multirow{2}*{\bf System} & \multicolumn{6}{c}{\centering \bf CMRC 2018} & \multicolumn{2}{c}{\centering \bf XNLI} & \bf TNEWS & \bf OCNLI & \multirow{2}*{\bf Average} \\
& \bf D-EM & \bf D-F1 & \bf T-EM & \bf T-F1  & \bf C-EM & \bf C-F1  & \bf Dev & \bf Test & \bf Dev & \bf Dev \\
\midrule
\bf PERT$_\text{base}$ (no limit) &  65.4 & 85.0 & 70.2 & 87.3 & 22.4 & 45.6 & 74.8 & 74.4 & 54.5 & 70.6 & 65.02 \\
┗Word 		& 59.3 & 80.6 & 64.8 & 83.5 & 12.6 & 32.2 & 73.2 & 72.1 & 53.2 & 69.3 & 60.08 \\
┗N-gram 	& 62.2 & 82.5 & 67.3 & 84.8 & 17.2 & 36.1 & 73.4 & 73.2 & 53.8 & 69.5 & 62.00 \\
┗Sentence 	& 63.7 & 83.2 & 69.1 & 86.2 & 16.8 & 38.0 & 74.3 & 73.0 & 54.1& 70.0 & 62.84 \\
\bottomrule
\end{tabular}
\end{center}
\caption{\label{analysis-different-granularity} Permutation within different granularities: word, N-gram, and sentence. We report five-run average for all results.}
\end{table*}

Among various permutation granularities, PERT with no permutation limit yields the best performances on all tasks.
We noticed that if we choose a smaller granularity (such as word), the system performance becomes the worst.
Though using a smaller granularity will make the input text more readable, it is less challenging to the pre-training task, and thus cannot extract useful semantics for text representation.

\subsection{Global v.s. Local Prediction}
The output space of PERT is the whole input sequence (the length of input), which is different from the MLM that predicts in the vocabulary space.
In this section, we modify the output of PERT to directly predict the original word in the vocabulary space instead of the original token's position.
We pre-train PERT$_\text{base}$ with 300K training steps.\footnote{Note that the hyperparameter for this PERT (as well as the one in Section \ref{sec-analysis-partial}) is slightly different from PERT$_\text{base}$ in Section \ref{sec-analysis-granularity}, and thus their baseline results are different.}
The results are shown in Table \ref{analysis-different-global}.
\begin{table*}[h]
\tiny
\begin{center}
\begin{tabular}{l c c c c c c | c c c c | c}
\toprule
\multirow{2}*{\bf System} & \multicolumn{6}{c}{\centering \bf CMRC 2018} & \multicolumn{2}{c}{\centering \bf XNLI} & \bf TNEWS & \bf OCNLI & \multirow{2}*{\bf Average} \\
& \bf D-EM & \bf D-F1 & \bf T-EM & \bf T-F1  & \bf C-EM & \bf C-F1  & \bf Dev & \bf Test & \bf Dev & \bf Dev \\
\midrule
\bf PERT$_\text{base}$ (local) 	& 64.1 & 84.0 & 69.1 & 86.5 & 21.0 & 43.3 & 74.1 & 74.4 & 54.5 & 70.6 & 64.16 \\
┗Global 				& 61.1 & 81.4 & 65.8 & 84.3 & 15.8 & 36.2 & 73.6 & 74.0 & 55.4 & 69.4 & 61.70 \\
┗Local + Global			& 63.2 & 83.6 & 67.7 & 85.8 & 19.3 & 42.0 & 74.6 & 74.6 & 55.1 & 70.0 & 63.59 \\
\bottomrule
\end{tabular}
\end{center}
\caption{\label{analysis-different-global} Comparison of local prediction (default) and global prediction.}
\end{table*}

The experimental results show that predicting in the vocabulary space is not necessary for PerLM, which is significantly worse than predicting on the local input sequence, where an average of 2.46 performance gap is observed.
Unfortunately, combining both local and glocal prediction (i.e., both predict the original token's position and the exact word in the vocabulary space) does not yield better performance than the local prediction only.
This reminds us that predicting the missing word in the global (vocabulary) space is not always necessary for pre-training tasks.

\subsection{Partial v.s. Full Prediction}\label{sec-analysis-partial}
Most of the pre-trained language model that adopts MLM-like pre-training task uses partial prediction.
The partial prediction only makes predictions on the masked tokens rather than the whole input sequence (full prediction).
Through the experiments in ELECTRA \citep{clark2020electra}, the authors present that the full prediction yields better performance than the partial prediction, and thus ELECTRA adopts full prediction for the discriminator using replaced token detection (RTD).
However, does it apply to other pre-training tasks other than RTD?
In this experiment, we compare the results on pre-training with full prediction and partial prediction.
We pre-train PERT$_\text{base}$ with 300K training steps.
The results are shown in Table \ref{analysis-different-partial}.
\begin{table*}[h]
\tiny
\begin{center}
\begin{tabular}{l c c c c c c | c c c c | c}
\toprule
\multirow{2}*{\bf System} & \multicolumn{6}{c}{\centering \bf CMRC 2018} & \multicolumn{2}{c}{\centering \bf XNLI} & \bf TNEWS & \bf OCNLI & \multirow{2}*{\bf Average} \\
& \bf D-EM & \bf D-F1 & \bf T-EM & \bf T-F1  & \bf C-EM & \bf C-F1  & \bf Dev & \bf Test & \bf Dev & \bf Dev \\
\midrule
\bf PERT$_\text{base}$ (partial) 	& 64.1 & 84.0 & 69.1 & 86.5 & 21.0 & 43.3 & 74.1 & 74.4 & 54.5 & 70.6 & 64.16 \\
┗Full Prediction 			& 63.7 & 84.1 & 68.0 & 86.1 & 18.7 & 40.9 & 74.3 & 73.9 & 54.2 & 71.0 & 63.49 \\
\bottomrule
\end{tabular}
\end{center}
\caption{\label{analysis-different-partial} Comparison of partial prediction (default) and full prediction.}
\end{table*}

As we can see that, full prediction does not yield better performance in PERT, where its performance is significantly worse in MRC tasks and yields similar performance in text classification tasks.
This demonstrates that it is not always effective to use full prediction in pre-training tasks  and should be adjusted to the nature of the designed pre-training task.

\section{Conclusion}
In this paper, we propose a new pre-trained language model, called PERT, which uses Permuted Language Model (PerLM) as the pre-training task.
The objective of PerLM is to predict the position of the original token in a shuffled input text, which is different from the MLM-like pre-training task.
To evaluate the performance of PERT, we carried out extensive experiments on both Chinese and English NLU tasks.
The experimental results show that PERT yields improvements on MRC and NER tasks but does not perform that well on TC tasks.
Additional quantitative analyses on PERT are also performed to better understand our model and the necessities of each design.
We hope that the trial of PERT can inspire our community to design non-MLM-like pre-training tasks for text representation learning.

\section*{Acknowledgments}\label{ack}
Yiming Cui would like to thank TPU Research Cloud (TRC) program for Cloud TPU access.

\bibliography{iclr2021_conference}
\bibliographystyle{iclr2021_conference}


\end{CJK*}
\end{document}